\documentclass[10pt, conference, compsocconf]{IEEEtran}
\usepackage{cite}

\ifCLASSINFOpdf
   \usepackage[pdftex]{graphicx}
\else
\fi
\usepackage[cmex10]{amsmath}
\usepackage{algorithmic}

\usepackage{array}

\usepackage{mdwmath}
\usepackage{mdwtab}

\usepackage{eqparbox}

\usepackage[tight,footnotesize]{subfigure}

\usepackage[font=footnotesize]{subfig}
\usepackage{stfloats}

\usepackage{url}

\begin{document}
%
\title{ A New Approach to Constraint Weight Learning for Variable Ordering in CSPs}


\author{\IEEEauthorblockN{Muhammad Rezaul Karim}
\IEEEauthorblockA{Department of Computer Science\\ 
University of Regina, Canada\\
Email: karim20m@uregina.ca}
}


%


\maketitle

\begin{abstract}
A Constraint Satisfaction Problem (CSP) is a framework used for modeling and solving constrained problems. Tree-search algorithms like backtracking  try to construct a solution to a CSP by selecting the variables of the problem one after another. The order in which these algorithm select the variables potentially have significant impact on the search performance. Various heuristics have been proposed for choosing good variable ordering. Many powerful variable ordering heuristics weigh the constraints first and then utilize the weights for selecting good order of the variables. Constraint weighting are basically employed to identify global bottlenecks in a CSP.

In this paper, we propose a new approach for learning weights for the constraints using competitive coevolutionary Genetic Algorithm (GA). Weights learned by the coevolutionary GA later help to make better choices for the first few variables in a search. In the competitive coevolutionary GA, constraints and candidate solutions for a CSP evolve together through an inverse fitness interaction process. We have conducted experiments on several random, quasi-random and patterned instances to measure the efficiency of the proposed approach. The results and analysis show that the proposed approach is good at learning weights to distinguish the hard constraints for quasi-random instances and forced satisfiable random instances generated with the Model  $RB$. For other type of instances, RNDI still seems to be the best approach as our experiments show.

\end{abstract}


%
\IEEEpeerreviewmaketitle
\section{Introduction}
Representing and solving problems involving constraints has important applications in artificial intelligence, including scheduling, planning, image interpretation and satisfiability testing. The idea of a static Constraint Satisfaction Problem (CSP) is to represent problem knowledge by defining constraints on the allowable values of problem variables. The basic algorithm to search solutions for a CSP is simple backtracking~\cite{Kumar92algorithmsfor}. In backtracking search, the basic operation is to pick one variable at a time, and consider one value for it at a time. The ordering in which the variables are labeled can affect the efficiency of the backtracking search~\cite{Dechter89experimentalevaluation,Gent96anempirical}. The variable ordering can affect the number of backtracks required in a search, which is one of the most important factors affecting the efficiency of an algorithm. When a lookahead strategy~\cite{Kumar92algorithmsfor} is incorporated with simple backtracking, the variable ordering can also affect the amount of search space pruned.

Dynamic or static variable ordering heuristics can be used to choose variables. In a static variable ordering (SVO) heuristic, variables are ordered before starting the search, and the search process always select variables in that order. Dynamic variable ordering (DVO) heuristics, on the other hand, select the current variable by extracting information during the search process. In~\cite{Dechter89experimentalevaluation,Zabih:1990} different SVO approaches have been proposed, while in~\cite{Bacchus95dynamicvariable,Smith97tryingharder,Gent96anempirical,refalo2004,Mouhoub:2011} different DVO approaches have been proposed. 

In this paper, we introduce a new approach where a competitive coevolutionary Genetic Algorithm (GA) is combined with the backtracking search to solve CSPs. Competitive co-evolution is a situation where two different species coevolve against each other. Typical examples of coevolving species are Predator-Prey and Host-Parasite. In this model, fitness of any individual from one species is determined through encounters with the individuals from the opposite population. In this paper, the coevolutionary GA is used to identify the hard constraints for a particular CSP by learning weights of all the constraints. Once the weights have been learned by the non-systematic coevolutionary GA, variables in the backtracking search is selected by a heuristic that uses weights to select a variable.

In order to evaluate the performance of the proposed approach, we have conducted a comparative experimental study with several similar existing variable ordering approaches. The experimental results based on thirty random, quasi-random and patterned instances show that the proposed approach is good at learning weights to distinguish the hard constraints for quasi-random instances and forced satisfiable random instances generated with the Model $RB$~\cite{Xu00exactphase}. For other type of instances, RNDI shows the best performance.

The rest of the paper is organized as follows: Section~\ref{csp} introduces CSP framework and solving methods for CSPs. Section~\ref{GA} provides an overview of GA, while section~\ref{relwork} discusses existing approaches for variable ordering in CSPs. The proposed approach for learning constraint weights using coevolutionary GA is detailed in section~\ref{learnweight}, followed by the report and analysis of the experimental results in section~\ref{section:exper}. The paper concludes in section~\ref{section:conclusions} with a summary of the work done and potential future work.

\section{CSPs}
\label{csp}
A typical CSP consists of a set of variables $V$, each with a domain $D_{i}$ of values, and a set of constraints $C$. Each constraint $x \in C$  is an arbitrary relation over a set of variables
and restricts the possible combinations of values of the associated variables. Many real world problems like scheduling problems, design problems,  workforce management, routing problems etc. can be modeled as CSPs. A solution to a CSP is an assignment of a value from its domain to every variable, in such a way that every constraint is satisfied. A CSP can have only one solution, more than one solution or no solution at all. A binary CSP (each constraint is either unary or binary) can be represented by a constraint graph~\cite{Kumar92algorithmsfor}. In a constraint graph, each node represents a variable, and each arc represents a constraint (relation) between variables represented by the end points of the arc. An arc representing unary constraint originates and terminates at the same node.

\subsection{Methods for Solving CSPs}
A CSP can be solved by systematic methods or  a non-systematic methods. A systematic method systematically explores the search space, while the latter does not. Backtracking search is a systematic method for solving a CSP. A backtracking search works by incrementally extending a partial solution to a complete solution. At each step of the backtracking search, the algorithm tries to assign a value to the the current variable and the attempt becomes successful if the assignment is consistent with the already assigned variables. If all the values in the domain of the current variable have been tried but no consistent assignment is not found, the algorithm backtracks to the preceding variable and try alternative values in its domain.

Standard backtracking has few limitations. The backtracking search can repeatedly fail due to the same reason which could be identified earlier in the search. This repeated failure is termed as {\em thrashing}. Local consistency techniques~\cite{Kumar92algorithmsfor}  have been proposed to overcome this difficulty. {\em Arc Consistency (AC)} is the most popular form of local consistency technique~\cite{Kumar92algorithmsfor}. Arc consistency eliminates values from domain of variables that can never be part of a consistent solution. An arc ($V_{i},V_{j}$) is arc consistent if  for all $ x \in D_{i}$, there exists a $y \in D_{j}$ such that $(x,y)$ is satisfied by the constraint. Arc consistency can be applied before the search or it can be integrated with the backtracking search. In the former case, AC can reduce the size of the search space before the search starts. In the latter case, AC helps to detect later failure earlier. Maintaining arc consistency throughout the search using an AC technique is called {\em MAC}~\cite{Sabin94MAC}. Each time a value is assigned to a variable, MAC algorithm enforces full arc consistency.

Non-systematic methods like local search and Evolutionary Algorithms~\cite{Goldberg1989} can also be used for tackling CSPs. These techniques are not guaranteed to find a solution, even if the CSP is consistent, as they rely on randomness. To use a non-systematic method, we have to find a suitable representation of the problem. We also need to define a problem specific fitness function to measure the quality of each potential solution.

\section{GA for Solving CSPs}
\label{GA}
GA is a stochastic population based global search and optimization method~\cite{Goldberg1989}. GA is a part of the group of Evolutionary Algorithms (EA)~\cite{Goldberg1989} and imitates the Darwinian evolution of the living beings. 

Like any EA, GA uses three main principles of the natural evolution: reproduction, natural selection and diversity of the species. GA maintains a population of potential solutions (chromosomes, strings or individuals). Initially binary chromosomes of individuals are created randomly. In a generation, some of the better individuals (parents) are selected based on a specific selection operator to generate offspring for the next generation. Genetic operators (crossover and/or mutation) are applied on the chromosomes of the selected candidates in a stochastic manner until a predefined number of offspring are created. Crossover is applied to two selected candidates to create one or two new candidate solutions. The purpose of crossover is to combine the good genetic material of the parents to create offspring with higher fitness. Mutation, on the other hand, is applied to one candidate and results in a new offspring.  The newly created offspring compete for a place in the next generation. Candidates (parents or offspring ) with higher fitness usually survive in the next generation. This process is iterated until a solution is found or a certain generation has been passed.
\subsection{GA Representation of CSPs}
GA can be used to represent CSPs primarily in two ways. One is standard integer-based  representation and the other is permutation based representation. The integer-based standard GA uses a string $S$ to represent a chromosome, where the $i^{th}$ element in $S$ corresponds to a value for variable $i$. For this type of representation, the fitness function usually measures the number of unsatisfied constraints. If constraints have weights, the sum of the weights of the unsatisfied constraints are used as fitness. The permutation representation, on the other hand, is based on a permutation of the variables of the CSP. A  decoder is used to transform a permutation to a partial instantiation by considering the variables in the order they occur in the chromosome. The decoder tries to assign the first possible domain value to the variable under consideration. If no consistent assignment is possible, the variable is penalized. For this type of representation, the total penalties are considered as the fitness of the chromosome (a permutation).

\section{Related Work}
\label{relwork}
Variable ordering heuristics can be classified into two categories: SVO and DVO. In a SVO heuristic, variables are ordered before the search starts. After that, variables are always selected in that order. {\em Smallest Domain First (SDF)}, is a SVO method in which variables are sorted in ascending order of domain size so that variables with smaller domain are instantiated first. {\em Maximum degree (deg)} is another example of SVO where the variable with the maximum degree in the constraint graph is chosen first~\cite{Dechter89experimentalevaluation}. Another example of SVO is {\em min bandwidth} which minimizes the bandwidth of the constraint graph~\cite{Zabih:1990}.

Mouhoub and Jafari~\cite{Mouhoub:2011} proposed two hybrid methods for variable ordering, where the variable ordering is decided before the start of the search. In the proposed approach, first a non-systematic approach based on Hill Climbing (HC) or Ant Colony Optimization (ACO) is applied to learn weights for constraints. After that, variables are sorted in descending order of weighted degree and the variables are instantiated in this order. The weighted degree ({\em wdeg}) of each variable equals to the sum of the weights of the constraints that the variable is involved in. The method based on Hill Climbing is referred to as HC/MAC in~\cite{Mouhoub:2011}, while the method based on ACO is referred to as ACO/MAC. The weight learning phase in  HC/MAC is an extension of the method proposed by Morris~\cite{Morris:1993}. Morris proposed that the weights of the violated constraints to be incremented when the local search enters into local minimum. But this method is not suitable for learning constraint weights to determine a good variable ordering. This is due to the reason that the search might take a long time to enter a local minimum. Mouhoub and Jafari extended this approach by proposing a cut off parameter. This parameter specifies the maximum number of iterations the local search should run before the next restart. In this approach,  search terminates each time it reaches the cut off point or enters a local minimum. Before the next search begins, the weights of violated constraints are incremented. In this approach, new search can be initiated upto only a certain number of time.

In contrast to SVO, DVO heuristics select the current variable using information that is made available during the search process. {\em dom}~\cite{Haralick79} is a very well known DVO heuristic which selects the variable that has the least remaining values in its domain. {\em ddeg} chooses the variables that are involved in the least amount of constraints with unassigned variables. {\em dom/ddeg}~\cite{Smith97tryingharder} is the combination of {\em dom} and {\em ddeg}. This heuristic selects a variable that has the minimum ratio of  {\em dom} to {\em ddeg}. Boussemart et al.  proposed a conflict driven variable ordering heuristics which uses MAC as the basic solving method~\cite{Boussemart04boostingsystematic}. In this method, during the constraint propagation phase, the weight of each constraint is incremented every time the constraint causes a domain wipe out. Whenever a variable needs to be selected, this technique selects the variable that has the largest weighted degree. This method has a potential limitation, this method might not have enough information about weights, when it needs to make most important choices, the first few variable selections. For this reason, size of the search space can significantly increase.

Grimes et al~\cite{Grimes2007} proposed two heuristics to improve the method proposed by Boussemart et al. These two techniques, known as Weighted Information gathering (WNDI) and RANDom Information gathering (RNDI), perform number of search restarts to gather information from different parts of the search space before starting the main search process. These restarts are used to learn weights for the constraints which help to make better choices for the first few variables. Experiments in~\cite{Balafoutis2008} show that RNDI is the better among the two heuristics. In RNDI, in the first R-1 runs, variables are selected randomly at variable selection points and a constraint weight is incremented when a constraint causes a domain wipeout. On the final restart, instead of random variable selection, {\em dom/wdeg} heuristic is used to select a variable. $dom/wdeg$ heuristic prefers a variable that has the least ratio of $dom$ and $wdeg$, where the former refers to the current domain size, while the latter refers to the current weighted degree. The constraints weights learned upto the R-1 runs are used in the last restart to make better decisons in the early stages of the main search.

\section{Variable Ordering Based on Constraint Weights}
\label{learnweight}
In this paper, we follow two steps to come up with a good variable ordering. In the first step, competitive coevolutionary GA is used to learn weights for all the constraints in a CSP. The main goal of this step is to identify the hard constraints in the CSP. In the next step, constraint weights are used to select variables in the backtracking search combined with constraint propagation.

\subsection{Coevolutionary GA for Learning Weights}
In the competitive coevolutionary model, two species evolve together through an inverse fitness interaction process. In this model, success (failure) of one species is considered as the failure (success) of the individuals of the other species. In nature, competitive coevolution represents a predator-prey complex. For survival, the prey always try to defend itself better from the predator. In response to that, the predator always try to improve its attacking strategies. 

In competitive coevolutionary computation, one species corresponds to the potential solutions for the problem in question, while the other species corresponds to certain tests a solution must satisfy~\cite{Paredis1994,Paredis95}. Competition between two individuals from the two populations is achieved through encounters. In an encounter, if the solution passes the test, the solution is rewarded while the test is penalized; if the solution fails, rewards are assigned in a reverse manner. Each individual in any population has to maintain a history of the results of encounters. The fitness of an individual is computed on this basis of the history of encounters. The fitness of a test or a solution is the sum of of its rewards or penalties in the history of its encounters.
\begin{figure}[t]
{\bf Algorithm} $ CoEvolutionaryGAWeightLearning$
  \begin{center}
\fontsize{8}{5}
{
\begin{algorithmic}[1]
\STATE $t \leftarrow 0$;
\STATE generate the initial population $P_{sol}(t)$ at random
\STATE evaluate individuals in $P_{sol}(t)$ with the {\em history} of encounters
\STATE evaluate individuals in $P_{cons}(t)$ with the {\em history} of encounters
\REPEAT
\FORALL{$i =1,2,..$  {\em no of encounters} }
\STATE select a solution from $P_{sol}(t)$;
\STATE select a constraint from $P_{cons}(t)$;
\STATE perform an {\em encounter} between the selected solution and the constraint
\STATE update {\em history} of the selected solution based on the encounter result
\STATE update {\em history} of the selected constraint based on the encounter result
\ENDFOR
\STATE select two parents from $P_{sol}(t)$ based on their current fitness
\STATE apply genetic operators on the parents to create one offspring
\STATE evaluate the offspring
\STATE Insert the newly created offspring into $P_{sol}(t)$
\STATE $t \leftarrow t + 1$
\UNTIL{termination criteria are satisfied}
\end{algorithmic}
\normalsize
}
 \end{center}
\caption[]{A competitive coevolutionary GA for learning constraint weight }
\label{fig:ECLifeCycle}
\vspace{-.3 cm}
\end{figure}

Figure~\ref{fig:ECLifeCycle} shows the competitive coevolutionary GA that we use for learning weights for constraints in a CSP. This algorithm is a modified verison of the competitive coevolutionary GA used in~\cite{Paredis1994}. This algorithm uses different reward/penalty scheme and selection strategies. In~\cite{Paredis1994}, coevolution was used to derive solutions for a CSP. In this paper, we use it for a different purpose to learn weights for the constraints in a CSP instance. Next, we describe the various features of the coevolutionary GA .

\subsubsection{Population}
In the co-evolutionary GA, there are two populations. The first population, $P_{sol}$, consists of GA representation of the possible solutions for a CSP, while the other population, $P_{cons}$, contains the constraints of the CSP. The first population evolves with the help of genetic operators. The constraint population, on the other hand, does not undergo any changes and is problem specific. We use the integer-based standard GA representation for the solution population.

\subsubsection{Initialization and Fitness Evaluation} 
We need to evaluate fitness for each randomly generated solution in the initial population, as well as the child generated by genetic operators. As described earlier, the fitness of the solution and the constraint population is computed by their achieved scores in the history of encounters. The {\em number of encounters} in a history is a predefined parameter. The fitness of a solution or a constraint is equal to the sum of all scores in their history of encounters. In an encounter, we assign a solution a score of 1 if it satisfies the constraint, while -1 if it does not satisfy the constraint. For the constraints inverse happens. The selected constraint is assigned a score of 1 if the solution does not satisfy the constraint, while -1 if it is satisfied. 

\subsubsection{Update History of Encounters} 
Unlike the standard GA, in competitive coevolutionary GA,  at the beginning of a generation, a predefined number of encounters between solutions and constraints are performed. The solutions and constraints are selected in such a way that encounter only takes place between fittest solutions and constraints. A selected constraint can prove its hardness to one of the best solutions, only if it is not satisfied by the selected solution.

\subsubsection{Selection and Variation Operators} 
We use {\em tournament} selection whenever we need to select a solution from the solution population. When an encounter takes place, we also need to select a constraint. In our approach,
a constraint is selected based on {\em linear ranking} selection. These selection schemes are used to emphasize the selection of best solutions and hard constraints. We employ one point crossover and binary mutation as primary variation operators for the solution population. As the constraint population does not go through evolution, no variation operator is required.

\subsubsection{Termination} 
The algorithm stops after a predefined number of generations. Once the algorithm stops, we discard the solution population and keep the constraint population. The fitness of each constraint is considered as the weight for that constraint.

\subsection{Variable Ordering}
Once we learn weights for all the constraints in a CSP, we use MAC as the basic solving method for the CSP. MAC maintains complete arc consistency throughout the search.  We implement the AC-3~\cite{Mackworth1977} for applying arc consistency in the MAC search. At each variable selection point, we use $wdeg$ heuristic that we discussed in section~\ref{relwork}, which prefers a variable that has the largest weighted degree ({\em wdeg}). To compute $wdeg$, the weighted degree of each variable, we use the weights for the constraints learned earlier through the coevolutionary GA. Like RNDI, we also increment a constraint weight when a constraint causes a domain wipeout in the MAC search.

\section{Experimentation}
\label{section:exper}
In this section, we report and analyze the results of our experiments for the proposed variable ordering heuristic. First, we introduce experimental settings. Then we compare the performance of the proposed approach with two other existing variable ordering heuristics: RNDI and HC/MAC. Both of these existing approaches depend on constraint weighting to achieve a variable ordering. 

\subsection{Experimental Settings}
In order to investigate the performance of the proposed method and compare with other relevant variable ordering heuristics, we performed experiments on 30 CSP instances taken from~\cite{XCSPbenchmark2013}. 12 instances were random instances generated with the Model D~\cite{Gent01randomconstraint} and Model RB~\cite{Xu2007}, 10 instances were quasi-random instances, while 8 instances were patterned instances. We implemented the HC/MAC and RNDI approaches to compare with the proposed approach. Our implementation of HC/MAC and RNDI uses {\em wdeg} heuristic in the MAC search and increments a constraint weight when a constraint causes a domain wipeout in the MAC search. We executed all experiments on a machine with AMD Athlon II CPU@2.8 GHz, 2 GB RAM and Ubuntu 12.04.3 operating system.

In the coevolutionary GA that we use for learning weights for the constraints in a CSP, we set history length to 10 and number of encounters in a generation to 20. We use 50 as the solution population size, while the constraint population size equals the number of constraints in a problem instance. We use one point crossover with crossover rate 0.7 or 0.9, while for mutation, we use bit mutation with mutation rate set to 0.01 or 0.2. For the linear ranking selection, we set the bias parameter to 2.0, while for the tournament selection, tournament size is set to 2. We use number of generations as a termination condition for the co-evolutionary GA. This termination condition is problem instance specific and we vary it from 2 to 15 generation. For each approach, a total of 50 runs are conducted for each test instance. 

RNDI has two parameters: R and C. The former is the number of restarts, while the latter is the maximum number of nodes that each of these restarts can visit. We test with different values of R  like 5, 25, 50, 100, 150 and 500. The value of the parameter C is set to ten times the number of variables in a problem instance. HC/MAC also has two parameters: the total number of iterations and cut-off. For easy problem instances, this approach needs less iterations, while for the harder instances it needs more iterations to explore the search space. For each instance, we try different values of iterations: 5, 10, 25, 50, 100 and 500. The cut-off parameter was set to 50 for all instances. For all approaches and all problem instances, timeout is set to 1200 seconds. 

\subsection{Description of the Instances}
The first set of random instances ({\em rand} series)  that we test in this paper is generated with Model $D$~\cite{Gent01randomconstraint} and situated at the phase transition of search~\cite{XCSPbenchmark2013,Xu00exactphase}. The instances that we use have different tightness values: 0.1, 0.2, 0. 35. The tightness $t$ denotes the probability that a pair of values is allowed by a relation. If $t$ is near 0 then the instance is likely to be very difficult, while a value near 1 indicates that it is easy to solve. The second set of random problems ({\em frb} series) that we test are random binary satisfiable CSP instances generated by the Model $RB$~\cite{Xu00exactphase, Xu2007,forcedCSP2009}. These instances are forced satisfiable instances. Each of these instances has a large number of variables and has tightness 0.25, which is also the exact phase transition point. Thus these instances are also hard to solve~\cite{forcedCSP2009}.

The geometric instances ({\em geo} series) are a kind of random instances (quasi-random). In these type of instances, a distance parameter is used such that $distance \leq sqrt(2) $. Two coordinates are then chosen in such a way that the associated point lies in the unit square and the selection process is random for each variable.  If the distance between any variable pair (x,y) is less than or equal to the distance parameter, then $(x,y)$ is added to the constraint graph as an arc. Another set of quasi-random instances that we test are the instances belonging to the {\em ehi} series. {\em ehi} instances are CSP instances which are converted from 3-SAT unsatisfiable instances using the dual method~\cite{Bacchus:2000}.  A 3-SAT instance is a SAT instance where each clause contains exactly 3 literals.

In the {\em Quasigroup Completion Problem (QCP)}, the goal is to determine whether there exists any way so that the remaining entries of the partial Latin square can be filled to obtain a complete Latin square~\cite{Gomes2002}. The {\em Balanced Quasigroup with Holes (QWH)} problem is a variant of the QCP. In this problem, instances are generated in such a way that they are guaranteed to be satisfiable~\cite{Gomes2002}. These problem instances are harder as the distribution of the holes is balanced. These two problems are represented by the series {\em bqwh} and {\em qcp}.  

\subsection{Performance Comparison}
In this section, we show the results of our experiments for all problem instances alongside the results of RNDI and HC/MAC algorithms. In our experiments, for each problem instance, we note the CPU time ($t$) for reaching the solution and the number of visited nodes by MAC algorithm ($n$). The CPU time is the sum of the time taken for learning weights and the time taken for the main MAC search. Even though we note $n$, our main goal is to minimize $t$. To compare two approaches, we determine the statistical significance of the differences in $t$ or $n$ using {\em Mann-Whitney U} test. If the p-value of the Mann-Whitney U test is less than 0.05, we assume that there is a significant difference between the two approaches compared. Mann-Whitney U test tells us whether two approaches are different but it cannot tell us how much one approach outperforms the another. We use {\em Vargha-Delaney A}~\cite{Delaney2000} measure for this purpose. The $A$ measure tells us the probability that one approach will achieve higher $t$ or $n$ than another approach. When the $A$ measure is above 0.5, the first approach outperforms the second one. When the $A$ measure is 0.5, the two approaches are equal. Otherwise, first approach performs worse than the second one. In our comparison, when we say one approach is better than the other on a particular instance, we mean that the difference observed between those two approaches in terms of  $t$ (or $n$) is statistically significant and the value of $A$ measure is above 0.5.

\subsubsection{Random Instances}
Table~\ref{tbl:expDmodel} and Table~\ref{tbl:expDmodelStat} show the experimental results for the random instances generated with the Model $D$. From our results, we notice that RNDI performs better than CoEvoGA/MAC and HC/MAC both in terms of $t$ and $n$. Among the seven tested {\em rand} series instances, RNDI is better than CoEvoGA/MAC for six instances in terms of the both performance metrics. Compared to HC/MAC, RNDI is better for two instances in terms of $t$, while in terms of the other metric, it is better only for a single instance. If we compare CoEvoGA/MAC to HC/MAC, we see that HC/MAC also performs better than CoEvoGA/MAC for two instances, both in terms of $t$ and $n$. From the results, we can conclude that RNDI is more suitable for hard random instances which are situated at the phase transition and generated with the Model $D$, while CoEvoGA/MAC has the worst performance for these type of instances.  

Our experiments with hard forced satisfiable random CSP instances generated with the Model $RB$ ({\em frb} series) show different behavior of these approaches. Table~\ref{tbl:expRBmodel} and~\ref{tbl:expRBmodelStat} show the results for those instances. Unlike the random instances generated with the Model $D$, CoEvoGA/MAC performs better than RNDI for two instances in terms of $t$. Only for one out of these two cases, the proposed approach is better than the latter approach in terms of the performance metric $n$.  Taking shorter time for the MAC search but longer time for overall CPU time ($t$), indicates that RNDI takes more time to learn weights for the constraints than the proposed CoEvoGA/MAC approach. If we compare RNDI with HC/MAC, we see that they show almost same level of performance.
 \begin{table*}[]  
 \caption {Average CPU time ($t$) for reaching a solution, as well as the average number of visited nodes ($n$) in the search tree. Average is computed over 50 runs. The results are for {\em random instances} generated with the Model D}
  \centering
  \label{tbl:expDmodel}
 \resizebox{0.71\textwidth}{!}{%
\begin{tabular}{rrrrrrrrr}
\hline
\textbf{Instance} &\multicolumn{2}{c}{\textbf{CoEvoGA/MAC}}&&\multicolumn{2}{c}{\textbf{RNDI}}& &\multicolumn{2}{c}{\textbf{HC/MAC}}\\
\cline{2-3} \cline{5-6} \cline{8-9} 
  & t & n& & t & n  & & t & n \\
\hline
\textbf{rand-2-40-8-753-100-0} &512.12  &87737.18 &  &377.6 & 62972.5 & & 480.76  & 52540.393  \\
\textbf{rand-2-40-8-753-100-1} &1099.3 & 179314.82 &  & 1050.5 & 173576.04 & & 1104.38 & 176517.56  \\
\textbf{rand-2-40-8-753-100-2} &596.36  & 116226.5 &  & 591.98 & 115898.26 & &597.24& 113793.12  \\
\textbf{rand-2-40-11-414-200-1} &1004.3  & 163788.32 &  &  874.86 & 147547.3 & & 961.18 & 157598.04  \\
\textbf{rand-2-40-11-414-200-2} & 991.26 & 217676.86 &  &  950.7 & 206582.98 & &  945.3 & 211702.62  \\
\textbf{rand-2-40-11-414-200-3} & 1026.06 & 176652.3 &  & 974.56  & 170487.68 & &  801.08 & 171634.86 \\
\textbf{rand-2-40-16-250-350-4} & 586.5 & 39189.9 &  &  87.1 & 5090.4 & & 380.3 & 27129.2 \\
\hline
\end{tabular} }
\vspace{-.3 cm}
 \end{table*}
  
 \begin{table*}[]  
 \caption {Statistical test results for {\em random instances} generated with the Model D. '*' indicates that the difference between the two specified methods as indicated by a
column is statistically significant. When there is statistically significant difference, {\em A} measure value is shown. When the $A$ measure is above 0.5, the first approach outperforms the second one, otherwise the second approach is better. '-' indicates that the difference is not statistically significant }
  \centering
  \label{tbl:expDmodelStat}
 \resizebox{0.75\textwidth}{!}{%
\begin{tabular}{rrrrrrrrr}
\hline
\textbf{Instance} &\multicolumn{2}{c}{\textbf{CoEvoGA/MAC Vs. RNDI}}&&\multicolumn{2}{c}{\textbf{CoEvoGA/MAC Vs. HC/MAC}}& &\multicolumn{2}{c}{\textbf{HC/MAC Vs. RNDI}}\\
\cline{2-3} \cline{5-6} \cline{8-9} 
  & t & n& & t & n  & & t & n \\
\hline
\textbf{rand-2-40-8-753-100-0} & 0.366(*) &  0.355(*) &  &  -  & -  & & -  & - \\
\textbf{rand-2-40-8-753-100-1} & 0.326(*) & 0.370(*) &  & - & - & & 0.309(*) & -  \\
\textbf{rand-2-40-8-753-100-2} & -  &- &  & - & - & & - & -  \\
\textbf{rand-2-40-11-414-200-1} & 0.114(*) & 0.182(*) &  & 0.617(*) & 0.398(*)& & 0.149(*) & 0.229(*) \\
\textbf{rand-2-40-11-414-200-2} & 0.446(*) & 0.289(*) &  &  0.315(*) & 0.376(*) & & - & -  \\
\textbf{rand-2-40-11-414-200-3} & 0.306(*) & 0.352(*) &  & 0.000(*)  & - & &0.999(*) & - \\
\textbf{rand-2-40-16-250-350-4} & 0.140(*) & 0.180(*) &  & - & - & &  - & - \\
\hline
\end{tabular} }
 \end{table*}
 \begin{table*}[]  
 \caption {Average CPU time ($t$) for reaching a solution, as well as the average number of visited nodes ($n$) in the search tree. Average is computed over 50 runs. The results are for forced satisfiable {\em random instances} generated with the Model RB}
  \centering
  \label{tbl:expRBmodel}
 \resizebox{0.71\textwidth}{!}{%
\begin{tabular}{rrrrrrrrr}
\hline
\textbf{Instance} &\multicolumn{2}{c}{\textbf{CoEvoGA/MAC}}&&\multicolumn{2}{c}{\textbf{RNDI}}& &\multicolumn{2}{c}{\textbf{HC/MAC}}\\
\cline{2-3} \cline{5-6} \cline{8-9} 
  & t & n& & t & n  & & t & n \\
\hline
\textbf{frb56-25-1} & 221.48  & 10112.88 &  & 230.84 & 9910.34 & & 374.66  & 16664.14  \\
\textbf{frb56-25-2} & 62.16 & 3132.6 &  & 101.76   & 3148.6  & & 57.2 & 2653.4 \\
\textbf{frb56-25-3} & 44.1  & 2249.86 &  & 78.12  & 2015.3 & & 54.36 & 2564.54  \\
\textbf{frb56-25-4} & 441.6 & 20378.34  &  & 290.74 & 9359.82 & & 290.66 & 13909.52  \\
\textbf{frb56-25-5} & 128.04 & 5860.42 &  &  160.06 & 5541.32 & & 118.34 & 5505.54  \\
\hline
\end{tabular} }
 \end{table*}
  
 \begin{table*}[]  
 \caption {Statistical test results for forced satisfiable {\em random instances} generated with the Model RB. '*' indicates that the difference between the two specified methods as indicated by a
column is statistically significant. When there is statistically significant difference, {\em A} measure value is also shown. When the $A$ measure is above 0.5, the first approach outperforms the second one, otherwise the second approach is better. '-' indicates that the difference is not statistically significant }
  \centering
  \label{tbl:expRBmodelStat}
 \resizebox{0.75\textwidth}{!}{%
\begin{tabular}{rrrrrrrrr}
\hline
\textbf{Instance} &\multicolumn{2}{c}{\textbf{CoEvoGA/MAC Vs. RNDI}}&&\multicolumn{2}{c}{\textbf{CoEvoGA/MAC Vs. HC/MAC}}& &\multicolumn{2}{c}{\textbf{HC/MAC Vs. RNDI}}\\
\cline{2-3} \cline{5-6} \cline{8-9} 
  & t & n& & t & n  & & t & n \\
\hline
\textbf{frb56-25-1} & - &  -  &  &  0.670(*) & -  & & 0.417(*) & 0.406(*)\\
 \textbf{frb56-25-2} & 0.728(*) & -&  &-& 0.345(*) & & 0.824(*) & -  \\
 \textbf{frb56-25-3} & 0.842(*) & - &  & - & -& & 0.843(*) & -  \\
 \textbf{frb56-25-4} & - & 0.298(*) &  & 0.413(*) & 0.4156(*) & & - & -  \\
\textbf{frb56-25-5} & - & - &  &  - & - & & 0.646(*) & - \\
\hline
\end{tabular} }
 \end{table*}
 
 \begin{table*}[t]  
 \caption {Average CPU time ($t$) for reaching a solution, as well as the average number of visited nodes ($n$) in the search tree. Average is computed over 50 runs. The results are for {\em quasi-random} instances}
  \centering
  \label{tbl:expQuasiRndom}
 \resizebox{0.71\textwidth}{!}{%
\begin{tabular}{rrrrrrrrr}
\hline
\textbf{Instance} &\multicolumn{2}{c}{\textbf{CoEvoGA/MAC}}&&\multicolumn{2}{c}{\textbf{RNDI}}& &\multicolumn{2}{c}{\textbf{HC/MAC}}\\
\cline{2-3} \cline{5-6} \cline{8-9} 
  & t & n& & t & n  & & t & n \\
\hline
\textbf{geo50-20-d4-75-10} & 84.26 & 3538.86 &  & 169.34 & 4567.0 & & 89.5  & 4140.7  \\
\textbf{geo50-20-d4-75-11} &149.52 & 5948.16 &  &123.84 & 4088.06 & &141.5  & 5661.5  \\
\textbf{geo50-20-d4-75-60} & 113.9 & 4031.7 &  & 109.92 & 2839.12 & & 120.06 & 4252.48  \\
\textbf{geo50-20-d4-75-61} & 1.68 & 128.54 &  & 34.62 & 5.48 & & 2.5 & 92.42  \\
  \textbf{geo50-20-d4-75-66} & 2.54  & 206.24 &  & 19.36 & 226.38  & & 3.1 & 146.24  \\
 \textbf{geo50-20-d4-75-67} & 151.84  & 6566.14 &  &  145.46 & 3311.88 & &156.82& 7135.76  \\
 \textbf{ehi-85-297-1 (unsat)} & 7.16  & 271.4 &  & 341.58  & 9.6 & &22.32& 244.44  \\
 \textbf{ehi-85-297-2 (unsat)} & 7.38  & 307.12 &  & 303.96  & 7.64  & & 22.22 & 264.44  \\
 \textbf{ehi-85-297-50 (unsat)} & 9.86  & 313.38 &  & 306.62  & 8.24 & & 23.02 & 275.66  \\
\textbf{ehi-85-297-51 (unsat)} & 11.54 & 446.24 &  & 10.29 & 381.58  & & 25.52 & 420.54 \\ 

\hline
\end{tabular} }
 \end{table*}
 
 \subsubsection{Quasi-Random Instances}
Table~\ref{tbl:expQuasiRndom}  and~\ref{tbl:expQuasiRndomStat} show the experimental results for the quasi-random instances: {\em geo} series instances and {\em ehi} series instances. On the {\em geo} instances, CoEvoGA/MAC performs statistically significantly better than RNDI for three instances in terms of $t$. RNDI fails to achieve better performance in terms of the same metric, but for three instances, it is better with respect to the other metric, $n$. CoEvoGA/MAC also performs better than HC/MAC for two instances in terms of $t$. 

If we compare the performance for the unsatisfiable {\em ehi} instances, we notice that CoEvoGA/MAC outperforms the other two approaches in terms of $t$. RNDI, on the other hand, performs better than the other approaches, in terms of $n$, but these improved performance comes at the expense of very bad performance on $t$. RNDI takes long time to learn the weights for the constraints, which affects the total CPU time taken ($t$). Our main goal here is to minimize the CPU time $t$. It seems that RNDI is not suitable for quasi-random instances. CoEvoGA/MAC shows promising results for the instances belonging to the tested quasi-random problems.

 \subsubsection{Patterned Instances}
If we compare the performance for patterned instances, we notice from Table~\ref{tbl:expPattern1} and~\ref{tbl:expPattern1Stat} that RNDI  is best suited to these type of instances. RNDI performs better than CoEvoGA/MAC for four instances. Its performance is also superior than HC/MAC for two instances in terms of $t$. For the instances, {\em bqwh-18-141-50} , {\em bqwh-18-141-99} and {\em qcp-15-120-0}, RNDI takes almost half the CPU time less than the time taken by other approaches. It also requires almost half the number of tree nodes less than the nodes visited by the other approaches. CoEvoGA/MAC, on the other hand, performs better than HC/MAC for two instances in terms of the both performance criteria. HC/MAC shows some instability in terms of $t$. The reason is that, for the instance, {\em qcp-15-120-2}, HC/MAC shows very poor performance and also it tends to exceed the time limit.
 
   \begin{table*}[t]  
 \caption { Statistical test results for {\em quasi-random} instances. '*' indicates that the difference between the two specified methods as indicated by a
column is statistically significant. When there is statistically significant difference, {\em A} measure value is also shown. When the $A$ measure is above 0.5, the first approach outperforms the second one, otherwise the second approach is better. '-' indicates that the difference is not statistically significant.}
  \centering
  \label{tbl:expQuasiRndomStat}
 \resizebox{0.75\textwidth}{!}{%
\begin{tabular}{rrrrrrrrr}
\hline
\textbf{Instance} &\multicolumn{2}{c}{\textbf{CoEvoGA/MAC Vs. RNDI}}&&\multicolumn{2}{c}{\textbf{CoEvoGA/MAC Vs. HC/MAC}}& &\multicolumn{2}{c}{\textbf{HC/MAC Vs. RNDI}}\\
\cline{2-3} \cline{5-6} \cline{8-9} 
  & t & n& & t & n  & & t & n \\
\hline
\textbf{geo50-20-d4-75-10} & 0.823(*) & 0.636(*) &  & - & - & & 0.798(*)  & -  \\
\textbf{geo50-20-d4-75-11} & - & 0.381(*) &  & - & -  & & -  & 0.343(*) \\
\textbf{geo50-20-d4-75-60} & - & 0.344(*) &  & - & - & & - & - \\
\textbf{geo50-20-d4-75-61} & 1.000(*) & - &  & 0.805(*) & - & & 1.000(*) & -  \\
  \textbf{geo50-20-d4-75-66} & 0.996(*) & - &  & 0.705(*) & 0.394(*) & & 1.000(*) & 0.589(*) \\
 \textbf{geo50-20-d4-75-67} & -  & 0.312(*) &  & - & - & &- & 0.231(*)  \\
  \textbf{ehi-85-297-1 (unsat)} & 1.000(*)  & 0.000(*) &  & 1.000(*)  & -  & & 1.000(*) & 0.018(*)  \\
 \textbf{ehi-85-297-2 (unsat)} & 1.000(*)  & 0.000(*) &  & 0.999(*)  & 0.380(*) & & 1.000(*) & 0.000(*)  \\
 \textbf{ehi-85-297-50 (unsat)}  & 1.000(*)  & 0.028(*) &  & 0.996(*)  & -  & & 1.000(*) & 0.000(*)  \\
\textbf{ehi-85-297-51 (unsat)}  & 1.000(*)  & 0.015(*) &  & 0.953(*)  & -  & & 1.000(*) & 0.000(*)  \\
\hline
\end{tabular} }
 \end{table*}
 
  \begin{table*}[]  
 \caption {Average CPU time ($t$) for reaching a solution, as well as the average number of visited nodes ($n$) in the search tree. Average is computed over 50 runs. The results are for {\em patterned} instances}
  \centering
  \label{tbl:expPattern1}
 \resizebox{0.71\textwidth}{!}{%
\begin{tabular}{rrrrrrrrr}
\hline
\textbf{Instance} &\multicolumn{2}{c}{\textbf{CoEvoGA/MAC}}&&\multicolumn{2}{c}{\textbf{RNDI}}& &\multicolumn{2}{c}{\textbf{HC/MAC}}\\
\cline{2-3} \cline{5-6} \cline{8-9} 
  & t & n& & t & n  & & t & n \\
\hline

\textbf{bqwh-18-141-47} &247 & 67686.18&  & 216.68  & 55972.02 & & 268.02 & 73332.54  \\
\textbf{bqwh-18-141-50} & 110.2 & 27963.66 &  & 49.88 & 9350.24 & & 104.14 & 26084.24 \\
 \textbf{bqwh-18-141-97} & 89.24 & 25804.42  &  & 124.68 & 33352.2  & & 151.14 & 41727.18  \\
  \textbf{bqwh-18-141-98} &36.54 & 9159.3 &  & 39.56 & 5998.56  & & 48.98 & 11143.68  \\
 \textbf{bqwh-18-141-99} & 234.78  & 66932.14 &  & 75.14 & 18339.48 & &146.94 & 42830.08  \\
  \textbf{qcp-15-120-0} & 402.7  & 42787.23  &  & 111.13 & 6415.63& & 318.2 & 34292.63 \\
 \textbf{qcp-15-120-1} & 108.33 & 9929.96 &  & 120.33  & 6352.76& & 137 & 11498.7 \\
 \textbf{qcp-15-120-2} & 374.36 & 46120.5 &  & 284.53 & 28694.96 & & 746.2& 92956.73 \\
\hline
\end{tabular} }
 \end{table*}
 
    \begin{table*}[]  
 \caption { Statistical test results for {\em patterned} instances. '*' indicates that the difference between the two specified methods as indicated by a
column is statistically significant. When there is statistically significant difference, {\em A} measure value is also shown. '-' indicates that the difference is not statistically significant. }
  \centering
  \label{tbl:expPattern1Stat}
 \resizebox{0.75\textwidth}{!}{%
\begin{tabular}{rrrrrrrrr}
\hline
\textbf{Instance} &\multicolumn{2}{c}{\textbf{CoEvoGA/MAC Vs. RNDI}}&&\multicolumn{2}{c}{\textbf{CoEvoGA/MAC Vs. HC/MAC}}& &\multicolumn{2}{c}{\textbf{HC/MAC Vs. RNDI}}\\
\cline{2-3} \cline{5-6} \cline{8-9} 
  & t & n& & t & n  & & t & n \\
\hline

\textbf{bqwh-18-141-47} & - & - &  & - & - & & - & -\\
\textbf{bqwh-18-141-50} & 0.340(*) & 0.224(*) &  & - & - & & - & 0.344(*) \\
 \textbf{bqwh-18-141-97} & 0.633(*) & - &  & 0.647(*) & 0.633(*) & & - & -  \\
   \textbf{bqwh-18-141-98} &- & - &  & - & -  & & - & 0.420(*) \\
 \textbf{bqwh-18-141-99} & 0.265(*)  & 0.300(*) &  & - & - & & 0.367(*) &0.292(*)  \\
 \textbf{qcp-15-120-0} & 0.403(*) & 0.286(*) &  & - & -& & - & 0.363(*)\\
 \textbf{qcp-15-120-1} & - & - &  & -  & -& & - & - \\
 \textbf{qcp-15-120-2} & - & - &  & 0.713(*) & 0.705(*) & & 0.228(*) & 0.183(*) \\
\hline
\end{tabular} }
\vspace{-.2cm}
 \end{table*}
\subsubsection{Discussion}
From our experiments, it seems that RNDI is the most suitable approach for hard random instances generated with the Model $D$, while for the hard forced satisfiable Model $RB$ instances, CoEvoGA/MAC seems to be the best choice. Among the other tested problem instances, CoEvoGA/MAC shows the best results for the quasi-random problem instances, while for the patterned instances, RNDI is the best. Our experimental results also show that, for some type of problems (e.g. quasi-random), RNDI spends long time to learn weights for the constraints which reduce the number of tree nodes visited in the MAC search, but increases the total CPU time taken. HC/MAC, on the other hand, shows some kind of instability as it tends to exceed the time limit for some patterned instances. Even though the proposed CoEvoGA/MAC approach cannot outperform the other approaches in all types of problems, it shows stability in terms of $t$ and $n$.

We also notice that as we increase the value of the parameter {\em number of iterations}, HC/MAC takes more time to learn weights for the constraints, which affects $t$. For easy instances, it should be set to a small value. For RNDI, the parameter {\em number of restarts} should be set to a small number to reduce the weights learning time, while for CoEvoGA/MAC, the number of generations should be set to a small value, for the same reason. It is worth noting that the preprocessing time taken by the coevolutionary GA can be further improved by parallelizing fitness evaluation of individuals, but the weight learning time for RNDI cannot be improved. For learning weights in RNDI, we need to perform several restarts. The weights learned by one restart, except the last one, is subsequently used by the next restart. This process cannot be benefited by parallelization due to inherent dependency.
\section{Conclusion}
\label{section:conclusions}
In this paper, we have introduced a new approach using coevolutionary GA to learn weights for the constraints, which can identify the global bottlenecks in a CSP. Like RNDI, weights learned by the coevolutionary GA, later help to make better choices for the first few variables, which are the most important choices in the search. Our experimental results on various problem instances show that the proposed approach is good at finding hard spots in the search space and performs better than the existing approaches on certain types of problems. In this paper, we have done experiments with a single population size and few values for crossover and mutation rate. In future, we will perform experiments on other values of these parameters to understand their effect on the performance of the proposed approach. Our experiments are conducted on three types of instances. In future, we will also look into other type of instances like real world instances. To reduce the weight learning time, we will also parallelize the fitness evaluation of individuals in the co-evolutionary GA using multi-core CPU.
\bibliographystyle{abbrv}

\end{document}